\documentclass[10pt,twocolumn,letterpaper]{article}

\usepackage{iccv}
\usepackage{times}
\usepackage{epsfig}
\usepackage{graphicx}
\usepackage{amsmath}
\usepackage{amssymb}
\usepackage[usenames,dvipsnames]{xcolor}
\usepackage{xspace}
\usepackage{subcaption}
\usepackage{booktabs}
\usepackage{multirow}
\usepackage{wrapfig}

\definecolor{supervisedgray}{gray}{0.5}

\def \modelname {Video OWL-ViT\xspace}

\usepackage[pagebackref=true,breaklinks=true,letterpaper=true,colorlinks,bookmarks=false]{hyperref}
\usepackage{cleveref}

\iccvfinalcopy %

\def\iccvPaperID{12571} %

\ificcvfinal\fi

\newcommand\blfootnote[1]{%
  \begingroup
  \renewcommand\thefootnote{}\footnote{#1}%
  \addtocounter{footnote}{-1}%
  \endgroup
}

\begin{document}

\title{\modelname: Temporally-consistent open-world localization in video}

\author{Georg Heigold\\
Google DeepMind\\
\and
Matthias Minderer\\
Google DeepMind\\
\and
Alexey Gritsenko\\
Google DeepMind\\
\and
Alex Bewley\\
Google DeepMind\\
\and
Daniel Keysers\\
Google DeepMind\\
\and
Mario Lučić\\
Google DeepMind\\
\and
Fisher Yu$^\dagger$\\
ETH Zurich\\
\and
Thomas Kipf\\
Google DeepMind\\
}

\maketitle
\ificcvfinal\thispagestyle{empty}\fi

\begin{abstract}
We present an architecture and a training recipe that adapts pre-trained open-world image models to localization in videos. Understanding the open visual world (without being constrained by fixed label spaces) is crucial for many real-world vision tasks. Contrastive pre-training on large image-text datasets has recently led to significant improvements for image-level tasks. For more structured tasks involving object localization applying pre-trained models is more challenging. This is particularly true for video tasks, where task-specific data is limited. We show successful transfer of open-world models by building on the OWL-ViT open-vocabulary detection model and adapting it to video by adding a transformer decoder. The decoder propagates object representations recurrently through time by using the output tokens for one frame as the object queries for the next. Our model is end-to-end trainable on video data and enjoys improved temporal consistency compared to tracking-by-detection baselines, while retaining the open-world capabilities of the backbone detector. We evaluate our model on the challenging TAO-OW benchmark and demonstrate that open-world capabilities, learned from large-scale image-text pre-training, can be transferred successfully to open-world localization across diverse videos.

\end{abstract}

\blfootnote{\kern-1.7em$^\dagger$Work done while at Google.\\ Correspondence: \href{mailto:heigold@google.com}{\texttt{heigold@google.com}}, \href{mailto:tkipf@google.com}{\texttt{tkipf@google.com}}}%

\section{Introduction}

A central goal in computer vision is to develop models that can understand diverse and novel scenarios in the visual world. While this has been difficult for methods developed on datasets with closed label spaces, web-scale image-text pretraining has recently led to dramatic improvements in open-world performance on a range of \emph{image-level} vision tasks~\cite{gu2021vild,minderer2022simple,kuo2023fvlm}.

However, challenges still remain for \emph{object-level} tasks on images and especially videos. First, object-level tasks require predicting more complex output structures compared to image-level tasks, making transfer of pretrained models more challenging. Second, training data for structured tasks is limited due to the prohibitive labeling cost. Therefore, a key research question is how to transfer the open-vocabulary capabilities of image-text models to object-level tasks like object detection and tracking.

For object detection, works such as ViLD~\cite{gu2021vild}, RegionCLIP~\cite{zhong2021regionclip}, OWL-ViT~\cite{minderer2022simple}, F-VLM~\cite{kuo2023fvlm} etc.\ demonstrate that image-level open-vocabulary capabilities can be transferred to object detection with relatively little detection-specific training data. Most recent works achieve this by combining image-text pre-trained encoder backbones (e.g.\ CLIP~\cite{radford2021clip}) with detection heads. By transferring semantic knowledge from the %
backbone, the resulting models are capable of detecting objects for which no localization annotations were present in the detection training data.

\begin{figure*}[tbp]
    \centering
    \includegraphics[width=0.83\textwidth,trim={0 17cm 7.5cm 0},clip]{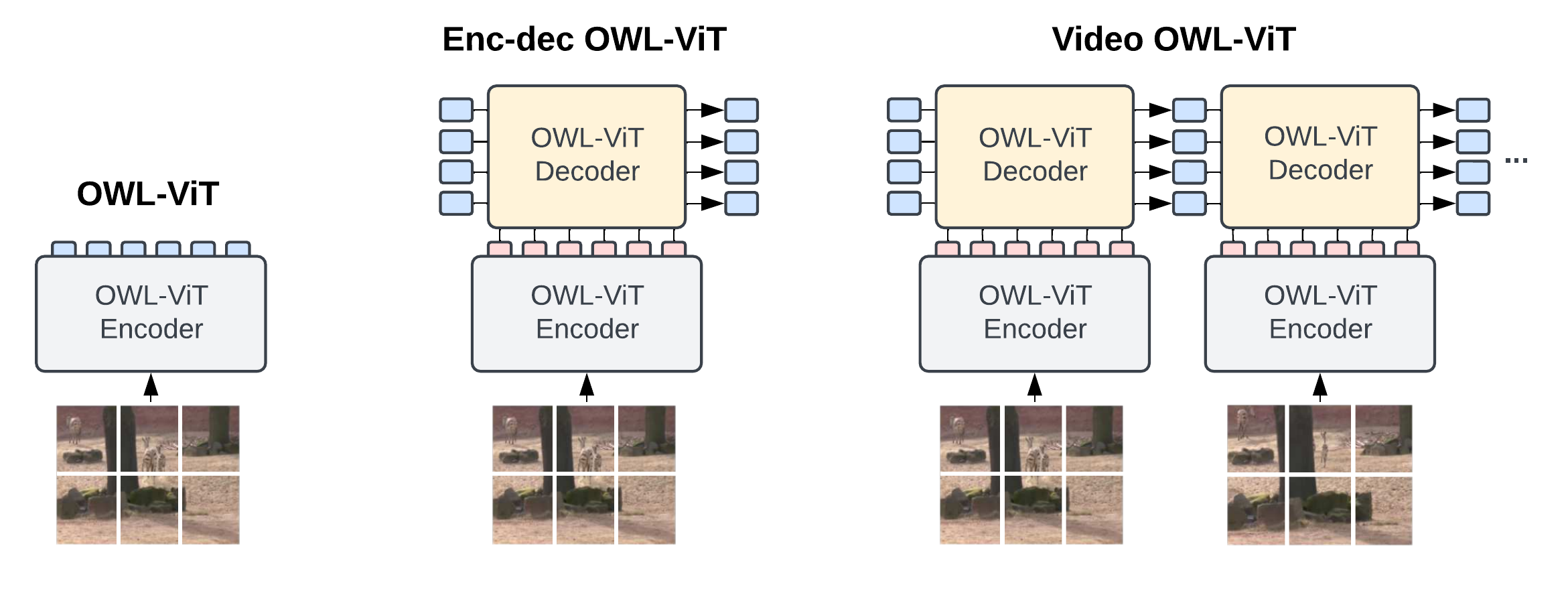}
    \vspace{-1em}
    \caption{\textbf{Model overview}. Our starting point is \textbf{OWL-ViT}~\cite{minderer2022simple} (\textit{left}), which uses an encoder-only Vision Transformer (ViT)~\cite{dosovitskiy2021vit} architecture for simple transfer from image-text pretraining to open-world detection: encoder tokens, arranged on the image grid, are used directly as object queries for detection. To transfer to temporal tasks without requiring frame-to-frame matching, we first develop a model variant inspired by DETR~\cite{carion2020detr} that decouples object queries from the image grid (\textbf{Enc-dec OWL-ViT}, \textit{middle}) by training a lightweight Transformer decoder on top of the ViT encoder while maintaining open-world detection capabilities. Finally, \textbf{Video OWL-ViT} (\textit{right}) simply connects the output of Enc-dec OWL-ViT applied to one frame to the next frame by using the predicted object queries as queries for the OWL-ViT Decoder of the next time step, without the need for any matching.}
    \label{fig:model}
\end{figure*}

Here, we extend this approach to video. We build on OWL-ViT~\cite{minderer2022simple}, which provides a simple open-world detection architecture in which light-weight box prediction and classification heads are trained on top of a CLIP backbone. We transfer the open-world capabilities of OWL-ViT to video understanding with minimal video-specific training data. The key idea behind our approach is to apply the open-world detector autoregressively to the frames of a video, propagating representations through time to track objects. To allow representations to bind consistently to the same object irrespective of its location, we depart from the encoder-only OWL-ViT architecture: We decouple object representations from the image grid by adding a transformer decoder, %
as is common practice in end-to-end closed-world detectors and trackers~\cite{carion2020detr,TrackFormer,yao2021efficient}. The decoder maps from image-centric encoder tokens to object-centric ``slots''. Information can then be carried through time by using the object slot representations from one frame as decoder queries on the next frame.

The object-centric decoder queries allow the model to \emph{learn} temporally consistent representations end-to-end from video data. This distinguishes our approach from previous open-world tracking models~\cite{Liu2021OpeningUO}, which applied frozen detectors frame-by-frame and used heuristics to link detections through time. 

We provide a recipe for incorporating a decoder into OWL-ViT and for fine-tuning the resulting model on video data without losing its open-world detection capabilities. We call the resulting model \emph{Video OWL-ViT}. We demonstrate strong performance on a challenging open-world video localization and tracking task, TAO-OW~\cite{Liu2021OpeningUO}, even for classes that were not seen during video training. We further demonstrate the zero-shot open-world generalization capabilities of Video OWL-ViT on a different dataset, YT-VIS~\cite{yt_vis}, that was not used for training. %

\section{Related Work}

\paragraph*{Open-Vocabulary Object Recognition}
Methods using pretraining on large amounts of web data, most notably 
Contrastive language-image pretraining (CLIP~\cite{radford2021clip}), have recently led to dramatic improvements in open-vocabulary performance on a range of vision tasks. Much research focuses on transferring these open-vocabulary capabilities %
to downstream tasks such as object detection.

The main challenge is to adapt the pretrained vision-language model to a downstream task in a way that retains the semantic knowledge and open-vocabulary capabilities acquired during pretraining. Various approaches have been proposed in the case of object detection, such as distillation~\cite{gu2021vild}, freezing the backbone~\cite{kuo2023fvlm}, or phrase grounding losses~\cite{li2021glip, zhang2022glipv2}. OWL-ViT~\cite{minderer2022simple} proposes a simple recipe to directly transfer a vision-language model to detection with minimal modifications. We build on OWL-ViT due to its simplicity and end-to-end architecture.

\paragraph*{Multiple Object Tracking (MOT)} The prevailing paradigm for MOT is tracking by detection, in which methods first locate the objects of interest and then associate detections across frames in a separate step. SORT~\cite{sort,deepsort} combined appearance cues with motion estimation and association optimization.
Many works have targeted %
better motion estimation~\cite{bergmann2019tracking,feichtenhofer2017detect,held2016learning,xiao2018simple}, while the across-frame association step can also be learned~\cite{braso2020learning,schulter2017deep}. 
Recently, many works~\cite{TrackFormer,zeng2022motr} explore Transformer~\cite{vaswani2017attention} encoder and decoder architectures to perform end-to-end learning of object tracking. Our work aims to extend this trend towards end-to-end open-world tracking.

\paragraph*{Open-World Tracking}
Recently, open-world tracking has been introduced as an extension of the MOT task~\cite{Liu2021OpeningUO,TETer}. In open-world tracking, the goal is to track not just known object categories, but all objects, including those from categories for which no annotated instances were seen during training. The ability to track unknown objects is critical for safety in applications such as autonomous driving. Since models have only recently become capable of strong open-vocabulary detection, few open-world tracking models exist. Baselines for open-world tracking obtain per-frame proposals from open-world detectors and use heuristics to link proposals over time~\cite{Liu2021OpeningUO,li2023ovtrack}. In concurrent work, OVTrack~\cite{li2023ovtrack}, Li et al.~similar to us propose data augmentation strategies to benefit from static images for learning temporal association, but they still rely on a tracking-by-detection heuristic to link detections over time. Here, we propose an end-to-end trainable architecture for open-world tracking.

\section{Method}
Our starting point is OWL-ViT~\cite{minderer2022simple} (Sec.~\ref{sec:owl_vit}), a simple yet effective Vision Transformer~\cite{dosovitskiy2021vit} model for open-world object detection in images. To generalize OWL-ViT to video tasks, we first develop an encoder-decoder variant of the model (Enc-dec OWL-ViT, Sec.~\ref{sec:enc_dec_owl_vit}) to decouple object queries from the image grid. This allows for a straightforward extension to video tasks, described in Sec.~\ref{sec:video_owl_vit} (Video OWL-ViT). For an overview of our method, see Figure \ref{fig:model}.

\subsection{Background: OWL-ViT}
\label{sec:owl_vit}

We briefly review the OWL-ViT model, which we use as our detection backbone.
OWL-ViT consists of a standard Vision Transformer~\cite{dosovitskiy2021vit} image encoder and an architecturally similar text encoder. The encoders are contrastively pretrained on a large datasets of image/text pairs~\cite{radford2021clip}. After pretraining, the model is transferred to detection by adding lightweight classification and box regression heads that predict class embeddings and box coordinates directly from the image encoder output tokens. 
For open-vocabulary classification, similarities are computed as the inner product between class embeddings derived from image patches and text embeddings of label names (provided text prompts). 
These similarities act as classification logits, which are shifted and scaled (using learned parameters), and trained using a sigmoid focal loss on standard detection datasets. To compute the loss, the Hungarian algorithm is used to match predictions to ground-truth targets. Unless otherwise noted, we use the CLIP-based L/14 variant of OWL-ViT. For detection training, we use the same data (Objects365~\cite{Objects365} and Visual Genome~\cite{krishnavisualgenome}) and augmentations as in the original paper~\cite{minderer2022simple}. Next, we describe how we adapt the model to tracking.

\subsection{Enc-Dec OWL-ViT}
\label{sec:enc_dec_owl_vit}

To enable temporally consistent representations that can track objects across frames, we must allow the model to decouple object representations from specific image tokens. We achieve this by inserting a Transformer decoder between the encoder and the object heads (i.e.~readout heads for box and class prediction), similar to the original DETR architecture~\cite{carion2020detr}. We use the decoder queries as ``slots'' that carry object representations recurrently from one timestep to the next. We refer to this architecture as Enc-dec OWL-ViT.

Given that OWL-ViT was originally designed as an encoder-only architecture, an important question is whether the open-vocabulary performance of the model can be maintained when adding a decoder. In \Cref{sec:results-upstream}, we show that the Enc-dec architecture retains most of the performance of the encoder-only architecture.

\begin{figure}[t!]
    \centering
    \includegraphics[width=0.95\linewidth,trim={0 21.5cm 29.5cm 0},clip]{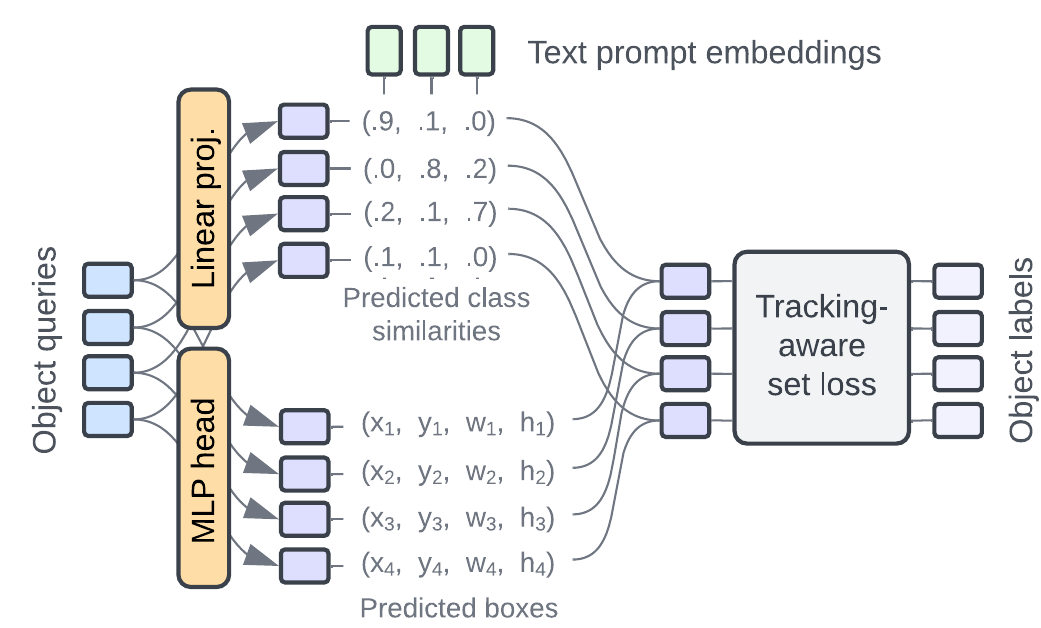}
    \caption{\textbf{Prediction heads and loss}. Video OWL-ViT predicts object bounding boxes and semantic embeddings from its set of object queries at every time step. Predicted semantic embeddings are compared against a set of text prompt embeddings to obtain predicted class similarity scores. For training, a tracking-aware matching-based set loss is used to compare predicted bounding boxes and class similarities against ground-truth object labels at every time step. %
    \label{fig:owl_vit_losses}}
\end{figure}

\subsection{Video OWL-ViT}
\label{sec:video_owl_vit}
We adapt the Enc-dec OWL-ViT model to video simply by repeatedly applying the model to frames of a video, one frame at a time, while using the predicted object queries of the previous frame as query initialization for the next time step. We further introduce video-specific data augmentation to make adaptation to video more label-efficient. %

\paragraph*{Architecture}
The architecture of Video OWL-ViT follows that of our image-based Enc-dec OWL-ViT model variant: To process a video, we apply Enc-dec OWL-ViT iteratively over the individual video frames. On the first video frame, we initialize object queries using simple learned feature vectors.
We then directly use the object queries predicted by the OWL-ViT decoder for this frame as decoder queries for the next time step. Model parameters are shared between time steps, i.e.\ the model is applied recurrently. 

We further carry over the box prediction and classification heads from upstream image-based pre-training. 
For fine-tuning, the image encoder and the text encoder are frozen (for efficiency), and only the box prediction and classification heads, and the transformer decoder, are updated. 

Finally, to obtain ``objectness'' scores for each instance at every frame, we take the maximum predicted classification logit across all classes.

\paragraph*{Training and loss} We train Video OWL-ViT using a tracking-aware set prediction loss similar to the tracking loss used in TrackFormer~\cite{TrackFormer}. 
In contrast to the TrackFormer setup, we train on short video sequences (instead of only pairs of frames). During training, we use Hungarian matching to match predicted object features (boxes and class similarities, see Figure~\ref{fig:owl_vit_losses}) to ground-truth object labels at every time step, starting from the first frame. Once a prediction is matched to a ground-truth track, it stays matched for the remainder of the training video clip, i.e.\ we only perform matching for previously unmatched objects. 

Video OWL-ViT predicts both class similarities (compared to text prompt embeddings) and object bounding box coordinates (center and width/height) at every time step. These predictions are used both for computing the matching cost as well as the final loss. Like OWL-ViT, we use focal sigmoid cross-entropy~\cite{zhu2020deformable} instead of the usual softmax cross-entropy, which is better suited for open-world detection, for both classification loss and matching cost. We use the same box prediction loss as in DETR~\cite{carion2020detr}, i.e.\ a weighted sum of L1 and generalized IoU loss terms. %

Object queries that are unmatched or matched to an empty track (with no object present in the ground-truth track at this time step) are trained to predict low class similarity scores for all provided text prompts, i.e.~all text prompts are treated as \textit{negative examples} in the loss. For matched object queries, we train the model to predict high class similarity scores for all ground-truth text prompts describing the class of that object and a low class similarity score for all other prompts, i.e.\ ground-truth text prompts are treated as \textit{positive examples} in the loss. %

Note that, when applied to a single frame, our tracking-aware set prediction loss exactly matches the loss used in OWL-ViT~\cite{minderer2022simple}.

\paragraph*{Augmentations}
As we primarily rely on upstream image-based pre-training and assume limited availability of video data, we make use of several video-specific data augmentation techniques.

\begin{figure*}[tbp]
    \centering
    \includegraphics[width=\linewidth,trim={2.6cm 8.5cm 0 8.5cm},clip]{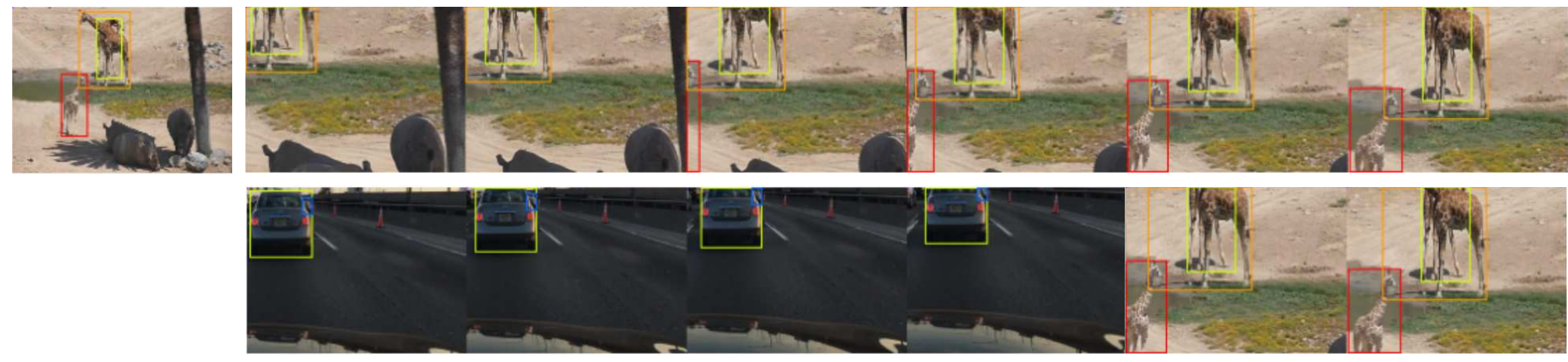}
    \caption{\textbf{Data augmentation}. Top row: Example of camera sweep with original image (left) and pseudo video (right). 
    Bottom row: Example of temporal mosaic (concatenation of two random clips).
    \label{fig:augmentations}}
\end{figure*}

First, we create \textit{pseudo videos} from images by aggressive scaling and cropping of an image, similar to prior work~\cite{TrackFormer,bridging-images-and-videos,TETer}, but with a linear motion model to generate video clips longer than 2 frames. This simulates a slowly moving camera observing a static scene (see Figure~\ref{fig:augmentations}, top row).

For video data, we perform temporally-consistent random re-scaling and cropping of entire video clips. To augment motion, we sub-sample the frame rate by randomly selecting frames while preserving temporal order.

We further find that extending the image mosaic augmentation used in OWL-ViT to video in the form of a \textit{temporal video mosaic} proves beneficial for generalization. This video mosaic augmentation is similar to VideoMix proposed in~\cite{Yun2020VideoMixRD} for video classification, and combines multiple video clips into a single clip by means of a scene cut (temporal video mosaic;  Figure~\ref{fig:augmentations}, bottom row).

Finally, for datasets like TAO that are only annotated at 1~FPS, we propagate annotations to non-annotated frames by linear interpolation to make use of all %
frames for training.

\section{Experiments}
Our experiments aim to answer the following main research questions:

\begin{itemize}
\addtolength{\itemsep}{-0.8ex}
    \item Does OWL-ViT maintain open-world detection performance when a decoder is added? 
    \item How well does the resulting model transfer to video (\textit{Video OWL-ViT})?
    \item Do open-world capabilities from image pre-training carry over to tracking of unseen instances in video?
\end{itemize}
We further perform detailed ablations to investigate individual components of our model.

\subsection{Encoder-Decoder Detection Evaluation}
\label{sec:results-upstream}
For our object detection backbone, we build on the CLIP-based OWL-ViT model~\cite{minderer2022simple} with a ViT-L/14 image encoder. To adapt this model for recurrent autoregressive application during tracking, we first investigate how detection performance is affected when adding a Transformer decoder between the image encoder and the prediction heads of the OWL-ViT architecture. 

Starting from the architecture in the original paper~\cite{minderer2022simple}, we pre-train the model on detection data (Objects365 and Visual Genome) for 70,000 steps with a batch size of 256 as described in the paper. We then add a Transformer decoder (architecture as in DETR~\cite{carion2020detr}; 6 layers, 8 heads, 4096 MLP dim, 1024 QKV dim, 100 decoder queries). To evaluate the effect of adding a decoder on detection performance, we train the Enc-dec model for an additional 70,000 steps using the same training data and schedule. We keep the image size at $672 \times 672$ for all experiments.

We find that detection performance on unseen LVIS ``rare'' classes ($\mathrm{AP}^\mathrm{LVIS}_\mathrm{rare}$) of the decoder model reaches 28.9, close to the 31.8 achieved by the encoder-only model. A small drop in performance is expected, given that the Enc-dec model outputs a significantly smaller number of object predictions (100 instead of 2304 for the encoder-only model). These results suggest that adding a decoder is a viable approach for adapting OWL-ViT to video.

\subsection{Open-world Video Localization}
\label{sec:exps-open-world-video-localization}

\paragraph*{Video training details} 
As in the previous section, we start from an Enc-dec OWL-ViT model in which the encoder is pretrained on detection and the decoder is randomly initialized. We found that pre-training the whole Enc-dec model on detection provided no advantage, likely because the Transformer decoder fulfils a different role in video compared to image data: it has to predict the objects in the current frame, but also produce suitable queries for the next frame to enable tracking. To simplify the experimental pipeline, we thus start video training directly from the original pre-trained OWL-ViT model. Video OWL-ViT uses a set of 196 learnable object queries.

We train for 100k steps with a batch size of 32 first on pseudo-videos only, followed by another 100k steps on a mixture of pseudo-videos and real videos from TAO-OW. Pseudo-videos are obtained by augmenting LVIS and Objects365 images as described in Section~\ref{sec:video_owl_vit}. Image resolution is kept at $672 \times 672$ by resizing images while preserving aspect ratios and padding as needed.

\paragraph*{Evaluation dataset}
We focus our evaluation on the recent TAO Open World (TAO-OW)~\cite{Liu2021OpeningUO} dataset: it is based on the Track Any Object (TAO) video dataset \cite{tao}, but specifically tests for open-world detection and tracking performance by (1)~restricting the object classes for which labels are provided during training, (2)~providing out-of-distribution validation and test sets to evaluate on unseen object classes, and (3)~introducing a metric that accounts for incomplete object annotations (e.g.\ due to filtering of known classes).

Since we are investigating the transfer of models pre-trained on large-scale web data, we restrict the set of ``seen'' object classes only during the final \emph{video} training stage, not during upstream image-level pre-training. Our only source of natural video training data is the TAO-OW training set, containing 500 videos.

As the test set annotations and evaluation server for TAO-OW are not yet publicly available, we evaluate our model on the provided validation set, containing 988 videos, and perform model selection and hyperparameter tuning based on the training set.

\begin{table*}[tbp]
\newcommand{\sv}[1]{\textcolor{supervisedgray}{#1}}
\caption{\textbf{TAO-OW} open world tracking. Baseline results from Liu et al.~\cite{Liu2021OpeningUO}.
Rows labeled ``w/o constraint'' do not use the non-overlapping constraint during evaluation. 
\sv{Gray} indicates results for classes that were seen during video training. AOA~\cite{Du20211stPS} performs video training on both known and unknown classes of TAO-OW and is thus not directly comparable (results marked in \sv{gray}). All metrics in $\%$. Best numbers highlighted in bold (excl.~results in \sv{gray}). %
}
\vspace{-1em}
\label{table:tao-ow-results}
\centering
\begin{tabular}{@{}lcccccc@{}} \\
\toprule
 & \multicolumn{3}{c}{\textbf{Known}}  & \multicolumn{3}{c}{\textbf{Unknown}} \\
 \cmidrule(l{2pt}r{2pt}){2-4} \cmidrule(l{2pt}r{2pt}){5-7}
\textbf{Model}                                  & OWTA$\uparrow$    & D.~Re.~$\uparrow$     & A.~Acc.~$\uparrow$    & OWTA$\uparrow$    & D.~Re.~$\uparrow$     & A.~Acc.~$\uparrow$ \\
\cmidrule(l{2pt}r{2pt}){1-4} \cmidrule(l{2pt}r{2pt}){5-7}
SORT \cite{sort}                                & \sv{46.6}         & \sv{67.4}             & \sv{33.7}                  & 33.9              & 43.4                  & 30.3\\
Tracktor \cite{bergmann2019tracking}            & \sv{57.9}         & \sv{80.2}             & \sv{42.6}                  & 22.8              & \textbf{54.0}                  & 10.0\\
OWTB \cite{Liu2021OpeningUO}                    & \sv{60.2}         & \sv{77.2}             & \sv{47.4}                  & 39.2              & 46.9                  & 34.5\\
OWL-ViT tracking-by-detection                   & \sv{37.7}         & \sv{36.1}             & \sv{40.1}                  & 31.0              & 32.0                  & 31.8\\
\textbf{Video OWL-ViT} (Ours)                   & \sv{59.0}         & \sv{69.0}             & \sv{51.5}                  & \textbf{45.4}              & 53.4                  & \textbf{40.5}\\
\cmidrule(l{2pt}r{2pt}){1-4} \cmidrule(l{2pt}r{2pt}){5-7}
AOA  (w/o constraint) \cite{Du20211stPS}        & \sv{52.8}         & \sv{72.5}             & \sv{39.1}                  & \sv{49.7}              & \sv{74.7}                  & \sv{33.4}\\
SORT-TAO  (w/o constraint) \cite{tao}           & \sv{54.2}         & \sv{74.0}             & \sv{40.6}                  & 39.9              & \textbf{68.8}                  & 24.1\\
OWTB (w/o constraint) \cite{Liu2021OpeningUO}   & \sv{60.8}         & \sv{82.0}             & \sv{45.5}                  & 42.4              & 58.9                  & 31.5\\
OWL-ViT tracking-by-det. (w/o constr.)     & \sv{44.5}         & \sv{45.5}             & \sv{43.9}                  & 42.2              & 51.5                  & 35.4\\
\textbf{Video OWL-ViT} (w/o constraint) (Ours)  & \sv{56.6}         & \sv{73.2}             & \sv{44.6}                  & \textbf{47.3}              & 62.3                  & \textbf{37.2}\\
\bottomrule
\end{tabular}
\end{table*}

\begin{figure*}[tbp]
    \centering
    \begin{subfigure}{0.17\linewidth}
    \includegraphics[angle=-90,width=\linewidth,trim={2cm 0cm 2cm 0cm},clip]{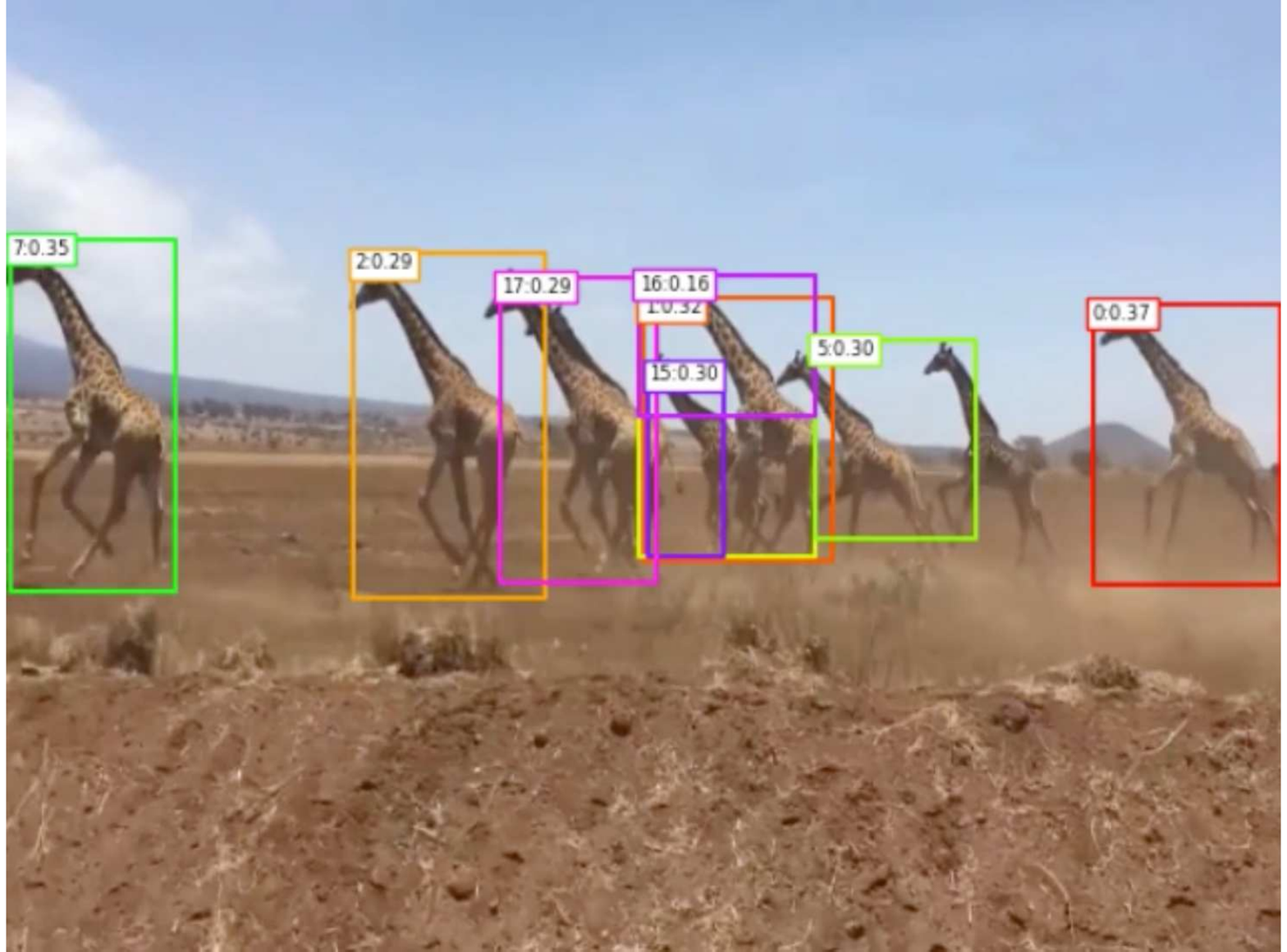}
    \end{subfigure}%
    \begin{subfigure}{0.17\linewidth}
    \includegraphics[angle=-90,width=\linewidth,trim={2cm 0cm 2cm 0cm},clip]{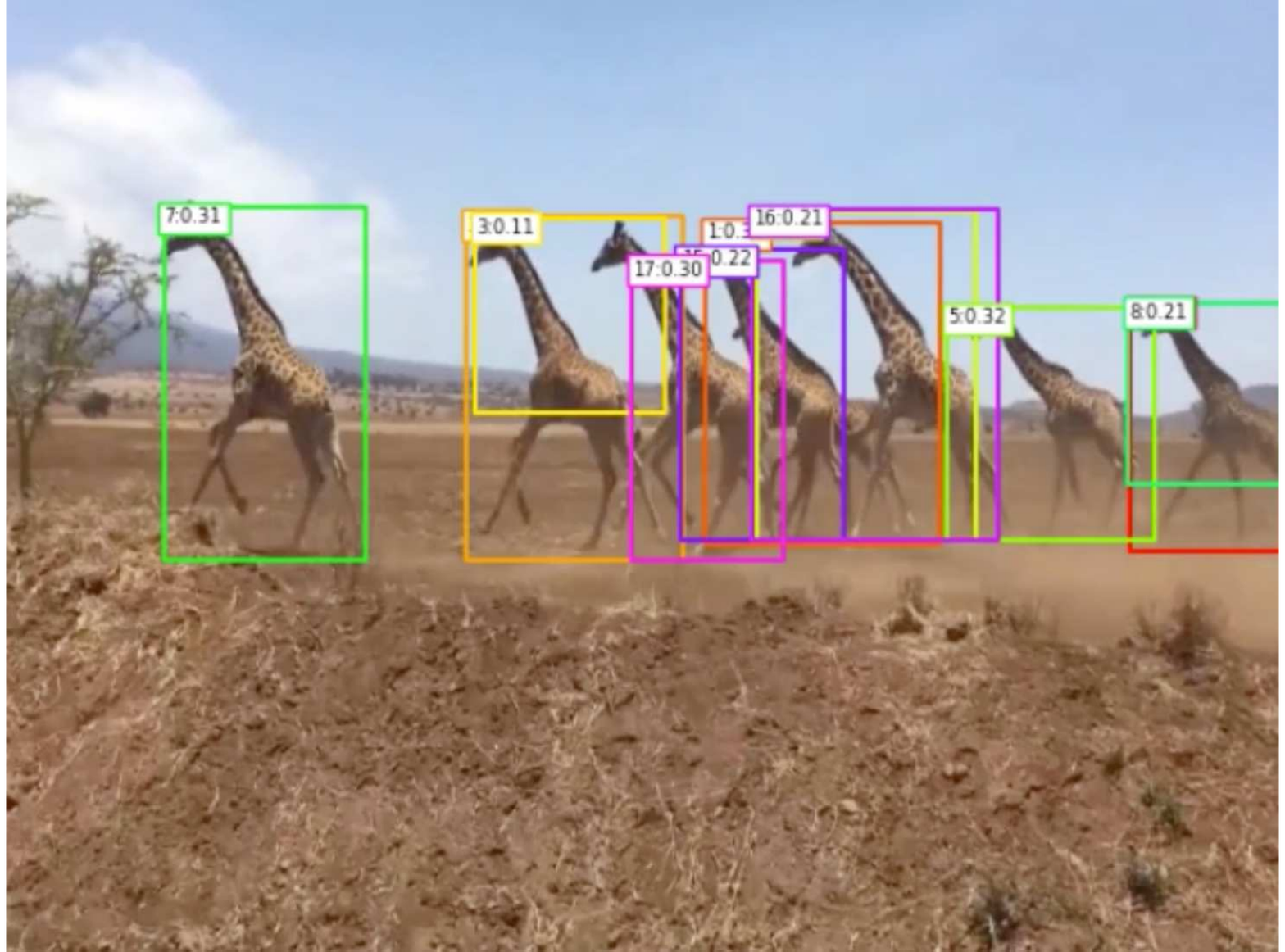}
    \end{subfigure}%
    \begin{subfigure}{0.17\linewidth}
    \includegraphics[angle=-90,width=\linewidth,trim={2cm 0cm 2cm 0cm},clip]{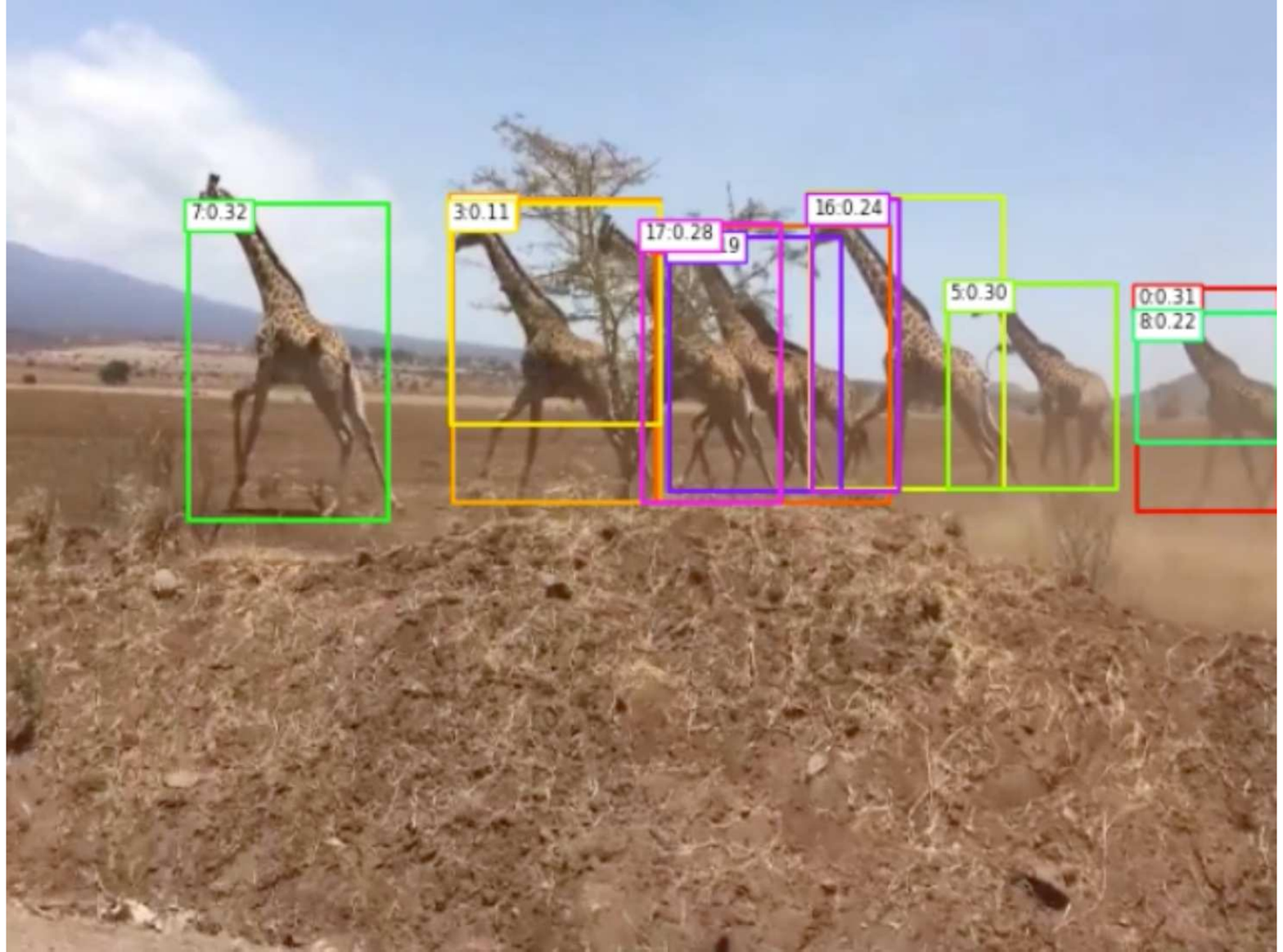}
    \end{subfigure}%
    \begin{subfigure}{0.17\linewidth}
    \includegraphics[angle=-90,width=\linewidth,trim={2cm 0cm 2cm 0cm},clip]{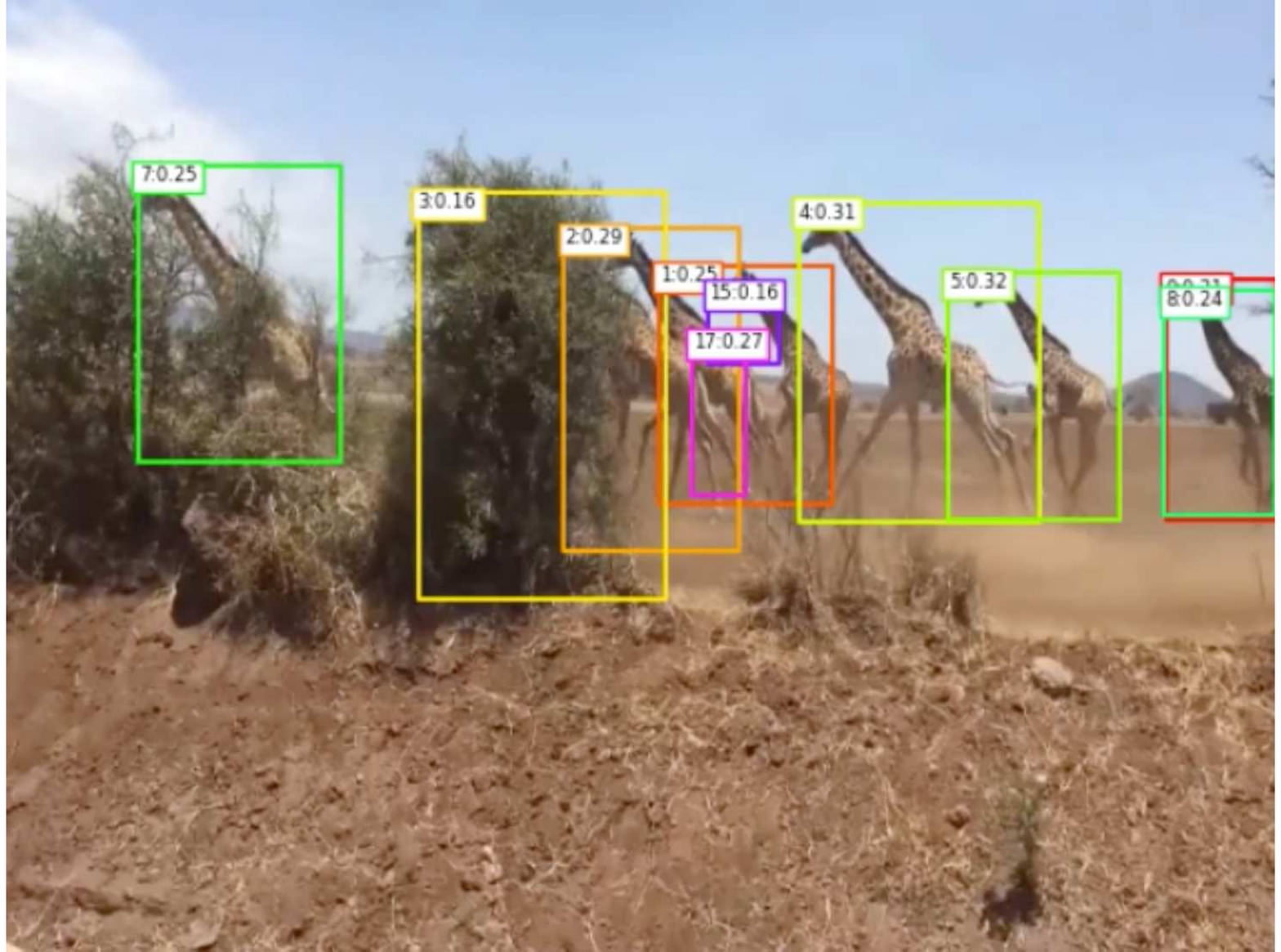}
    \end{subfigure}%
    \begin{subfigure}{0.17\linewidth}
    \includegraphics[angle=-90,width=\linewidth,trim={2cm 0cm 2cm 0cm},clip]{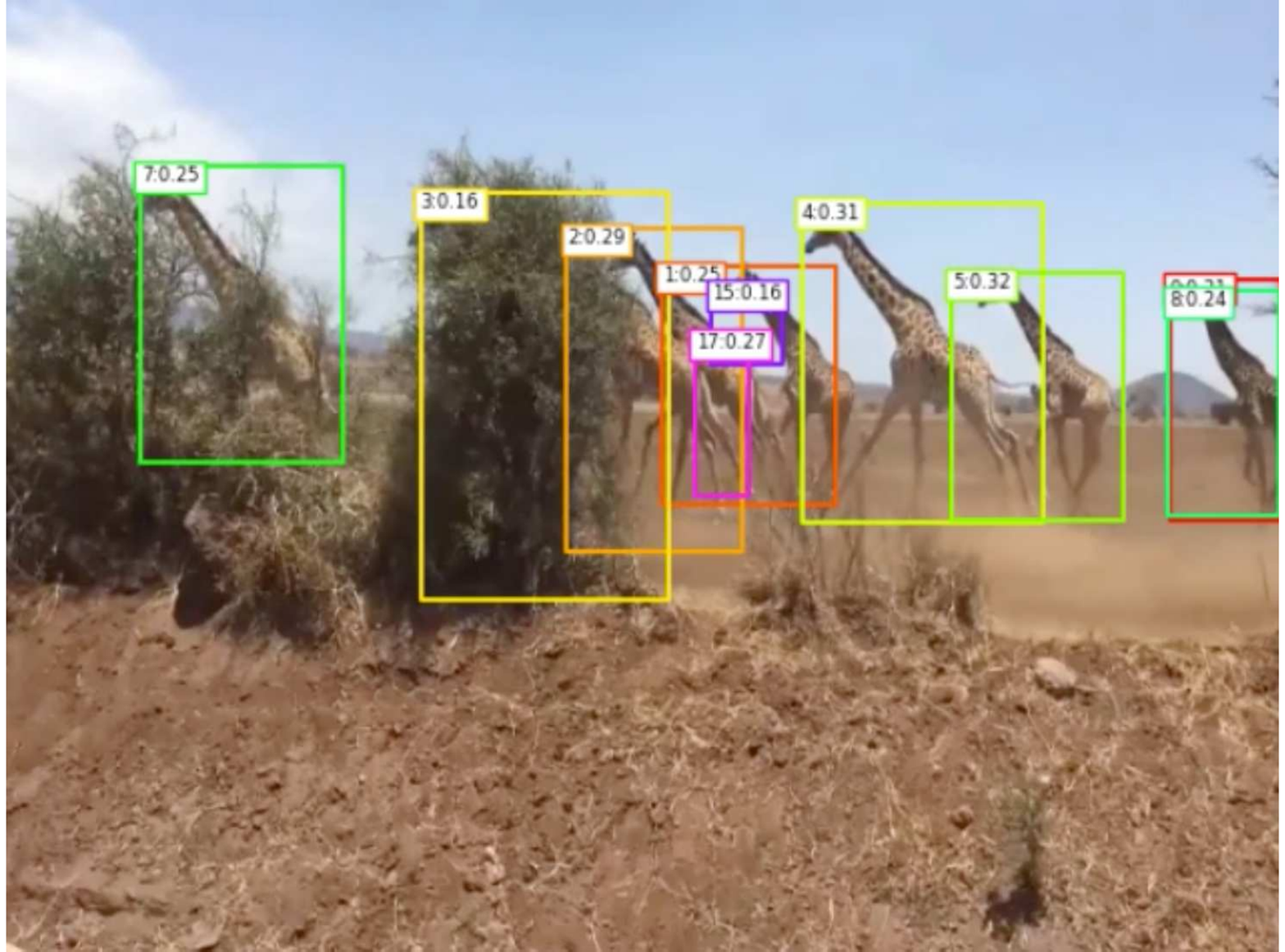}
    \end{subfigure}%
    \begin{subfigure}{0.17\linewidth}
    \includegraphics[angle=-90,width=\linewidth,trim={2cm 0cm 2cm 0cm},clip]{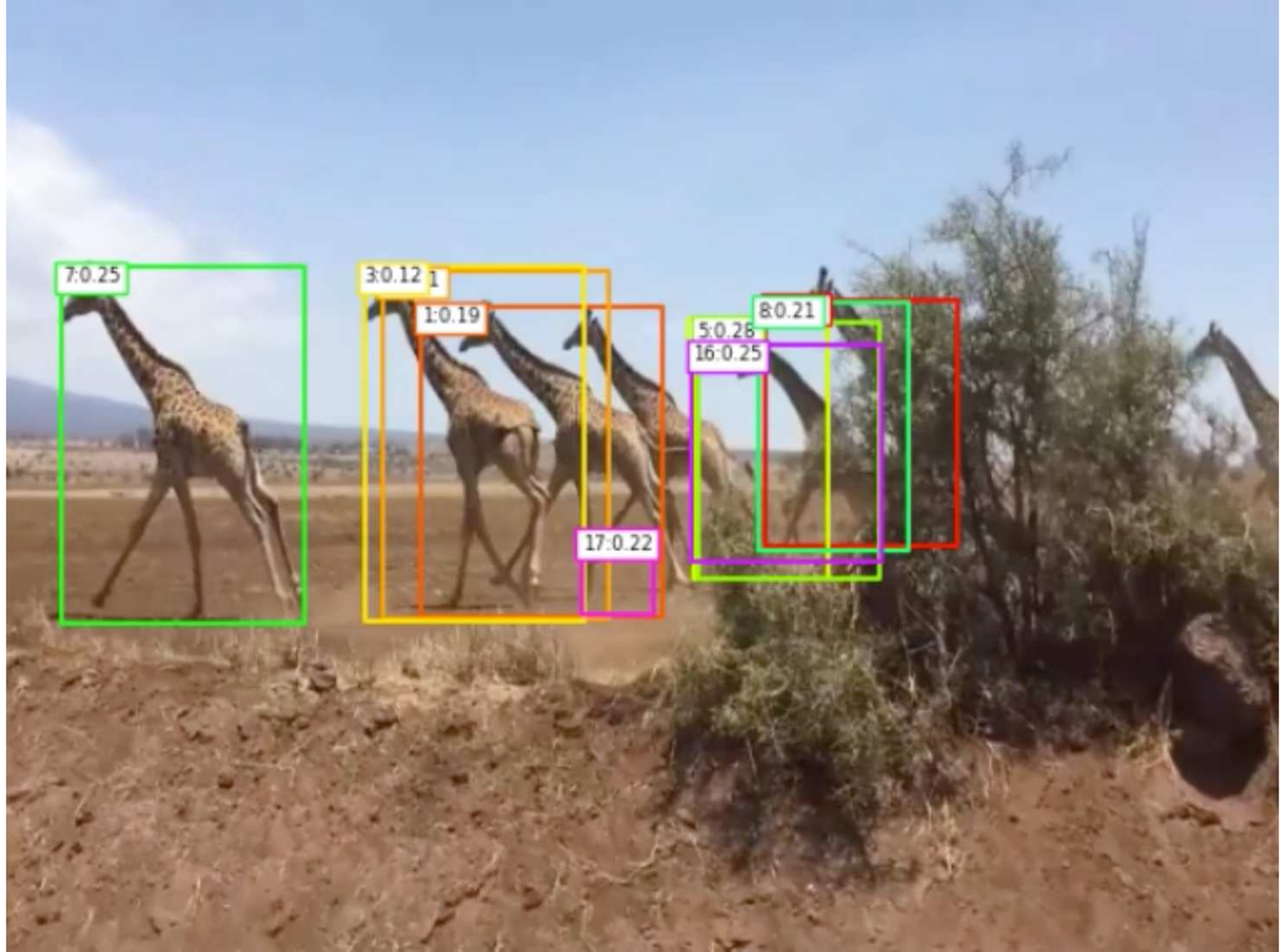}
    \end{subfigure}
    \caption{Qualitative example for Video OWL-ViT detection and tracking of multiple instances on the \textbf{TAO-OW} validation set. This example includes different instances of the same kind (giraffes) and partial occlusion. Video OWL-ViT can recover from occlusion, despite being trained on short video clips. \label{fig:tao-ow-examples}}
\end{figure*}

\paragraph*{Evaluation metric} For evaluation and model comparison, we use the Open-World Tracking Accuracy (OWTA) metric introduced by Liu et al.~\cite{Liu2021OpeningUO}, the standard metric for TAO-OW. OWTA is defined as the geometric mean of Detection Recall (D.~Re.) and frame-to-frame Association Accuracy (A.~Acc.), integrated over multiple localization thresholds. 
Importantly, D.~Re.~ignores false positive detections and is thus suitable for evaluating in the incompletely annotated open-world setting of TAO-OW. To avoid cheating the metric by producing a myriad of detection proposals, OWTA enforces a non-overlapping prediction constraint by requiring models to produce non-overlapping segmentation masks, so that every pixel is assigned to at most one detected object, or to the background.

To enforce this constraint, we train a separate segmentation head on top of the frozen OWL-ViT model (for details see appendix).  %
Specifically, for each frame and pixel, among all instances whose mask overlaps with that pixel, we keep the instance with the highest score. Pixels belonging to other instances are removed from the segmentation mask. The remaining instance mask is then converted back to box coordinates for evaluation. 

To account for the observation that predicted objectness scores can be miscalibrated for small objects and short tracks, we introduce two simple heuristics. To rank instances for overlap removal, we use the objectness score divided by the box area. This heuristic accounts for the observation that smaller objects tend to have lower objectness scores.
To suppress false positives for short tracks, we mark parts of tracks as background if they have significantly lower objectness scores than the maximum objectness observed in the track. We mark any detection as background if its objectness score is lower than a pre-defined fraction of the maximum objectness score along the track.

\paragraph*{Main results}

Our method makes two main contributions over heuristic tracking-by-detection baselines such as the Open World Tracking Baseline (OWTB)~\cite{Liu2021OpeningUO}: (1)~Our model is end-to-end trainable on video and can therefore learn temporal consistency directly from data. (2)~Our method transfers open-world semantic knowledge from image pretraining to tracking. Our results on TAO-OW show that both contributions translate to improved performance.

The advantage of end-to-end training is apparent from the association accuracy (A.~Acc., \Cref{table:tao-ow-results}), which measures the accuracy of associating detections across frames. Our model outperforms all baselines on A.~Acc.\ on both know and unknown classes. We hypothesize that learning temporal associations from data, rather than matching single-frame detections heuristically, reduces tracking error accumulation. An end-to-end method like ours also promises to improve further when more video training data is available. We provide qualitative results in Figure~\ref{fig:tao-ow-examples} and in video format in the supplementary material.

The transfer of pretrained knowledge is apparent from the performance of Video OWL-ViT on unknown classes, i.e.\ classes for which no video training data is available. Video OWL-ViT outperforms OWTB on unknown classes on the OWTA metric by a wide margin, showing that image-level open-world knowledge can be transferred to video with minimal video-specific training data. While we observe variance of OWTA scores on unknown classes between repeated training runs of approx.~1\% (absolute) OWTA, Video OWL-ViT still reliably outperforms the baselines.

On known classes, Video OWL-ViT performs similarly to OWTB on the overall OWTA metric. However, while the baseline achieves its performance primarily through high detection recall (D.~Re.\ in \Cref{table:tao-ow-results}, i.e.\ single-frame performance) which compensates for its low association accuracy, the end-to-end Video OWL-ViT has more balanced detection and association performance.

We compare models both with and without the non-overlapping constraint of the OWTA metric. For evaluations without the constraint, we note that our model uses significantly fewer tracks, a total of 196, compared to the most competitive baseline (OWTB), which uses the top 1000 object proposals per frame, thus placing our model at a disadvantage in this constraint-free evaluation: a higher proposal budget makes it easier to score well on detection recall. Despite this disadvantage, Video OWL-ViT compares favorably against the baselines reported previously on this benchmark.
\\[1em]
To isolate the benefit of the Video OWL-ViT decoder we also compare to a simple tracking-by-detection baseline based on appearance matching, depicted as \emph{OWL-ViT tracking-by-detection} in \Cref{table:tao-ow-results}. Specifically, our baseline performs optimal bipartite matching using the cosine similarity of the OWL-ViT embeddings from 300 proposals in the current and previous frame using the Jonker-Volgenant algorithm~\cite{jonker1988shortest}. These matches are extended to form tracks across the entire sequence. For unknown classes, this baseline performs comparably to the OWTB without constraint, with a bias towards better association. Video OWL-ViT shows a substantial gain over this baseline, highlighting the benefit of end-to-end training enabled by our model. In terms of computational cost, we find that our temporal decoder adds approx.~$6$ ms to the to the inference time, which is similar to the bipartite matching in our tracking-by-detection baseline ($4$--$11$ ms). %

\paragraph*{Zero-shot transfer}
We study generalization capabilities of Video OWL-ViT by evaluating the model without any further training on the YT-VIS 2019 dataset~\cite{yt_vis}, which contains annotations at significantly higher frame rates. The annotations for the official validation and test datasets of YT-VIS~\cite{yt_vis} are not published. For this reason we use validation and test splits taken from the training data with 200 videos each.\footnote{\url{https://www.tensorflow.org/datasets/catalog/youtube_vis}; Version:  `480\_640\_only\_frames\_with\_labels\_train\_split'} %
We report results on the test split.
In this zero-shot transfer setup, we measure the same metrics as used for TAO-OW, but evaluate at a higher frame rate of 6 FPS. We distinguish between known and unknown classes based on which classes were available in the TAO-OW training set.

\begin{figure*}[htp!]
    \centering
    \begin{subfigure}{0.49\linewidth}
    \includegraphics[angle=-90,width=\linewidth,trim={7cm 0cm 7cm 0cm},clip]{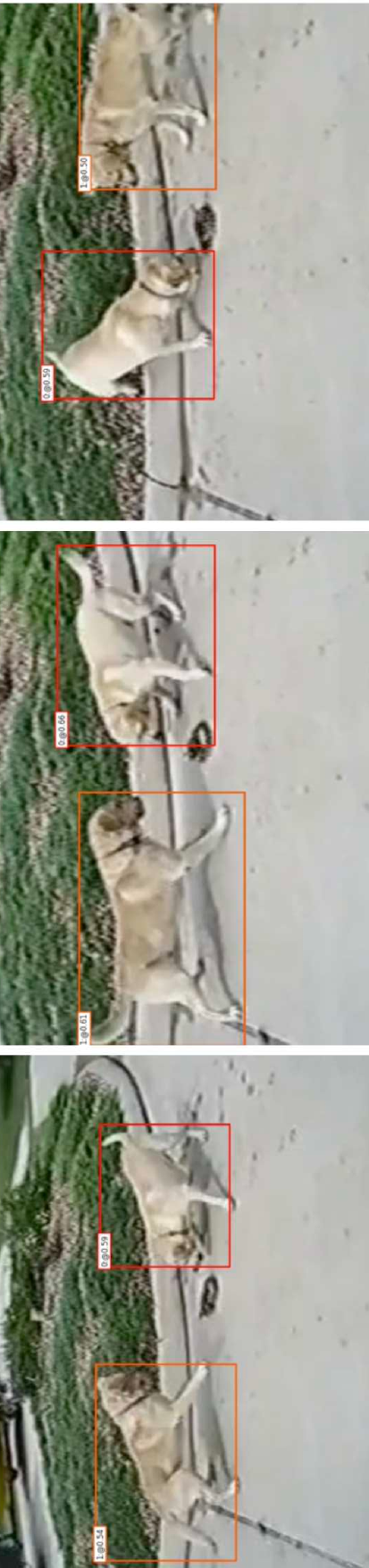}
    \end{subfigure}%
    \hfill
    \begin{subfigure}{0.49\linewidth}
    \includegraphics[angle=-90,width=\linewidth,trim={7cm 0cm 7cm 0cm},clip]{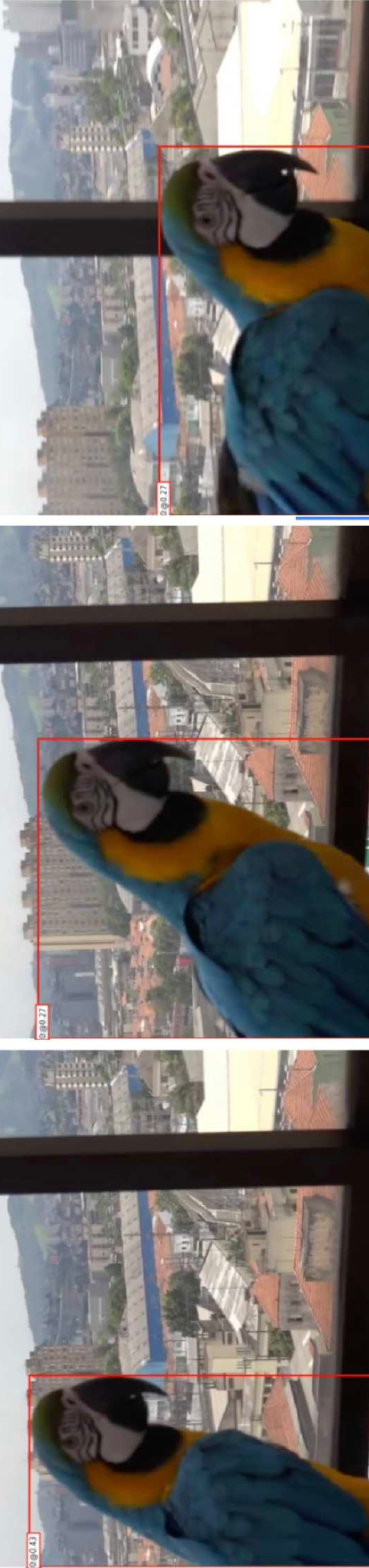}
    \end{subfigure}
    \caption{Zero-shot transfer to \textbf{YT-VIS} open world tracking with Video OWL-ViT. 
    Examples for a known class (``dog'', left) and an unknown class (``parrot'', right). The 6-second videos are visualized by three equidistant frames each.
    \label{fig:yt_vis-ow-examples}}
\end{figure*}

As can be seen in Table~\ref{table:yt_vis-ow-results} (quantitative results) and Figure~\ref{fig:yt_vis-ow-examples} (qualitative results), Video OWL-ViT shows strong transfer even to unseen classes in YT-VIS, despite evaluation at a significantly higher frame-rate. Compared to the \textit{OWL-ViT tracking-by-detection (TbD)} and \textit{SORT}~\cite{sort} baselines, Video OWL-ViT generally shows better transfer to both known and unknown classes, and demonstrates improved association accuracy.%

\begin{table}[tbp]
\caption{Zero-shot transfer to \textbf{YT-VIS} open world tracking with Video OWL-ViT. All metrics in $\%$.}%
\vspace{-1em}
\label{table:yt_vis-ow-results}
\centering
\resizebox{\linewidth}{!}{
\addtolength{\tabcolsep}{-2pt}
\begin{tabular}{@{}lcccccc@{}} \\
\toprule
 & \multicolumn{3}{c}{\textbf{Known}}  & \multicolumn{3}{c}{\textbf{Unknown}} \\
 \cmidrule(l{2pt}r{2pt}){2-4} \cmidrule(l{2pt}r{2pt}){5-7}
\textbf{Model}  & OWTA & D.~Re. & A.~Acc. & OWTA & D.~Re. & A.~Acc. \\
\cmidrule(l{2pt}r{2pt}){1-4} \cmidrule(l{2pt}r{2pt}){5-7}
SORT~\cite{sort} & 43.0 & 40.5 & 48.7 & 45.0 & 48.1 & 44.9\\
OWL-ViT TbD & 71.1 & 72.3 & 70.8 & 71.9 & 73.5 & 71.5\\
\textbf{Video OWL-ViT} & 79.4 & 82.9 & 76.4 & 81.2 & 81.8 & 81.0\\

\bottomrule
\end{tabular}
}
\end{table}

\begin{figure}[htp!]
    \centering
    \vspace{-1mm}
    \includegraphics[width=0.9\linewidth]{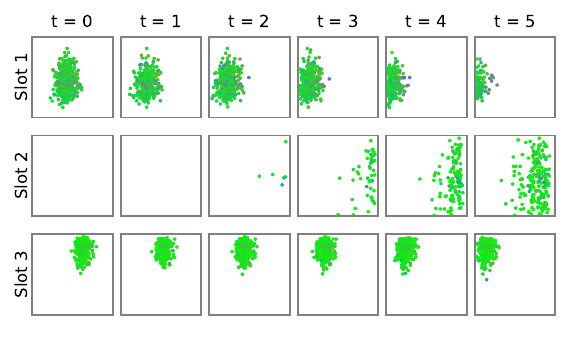}
    \vspace{-2mm}
    \caption{Visualization of box predictions on all images from the COCO 2017 val set, similar to Figure~7 in DETR~\cite{carion2020end}.
    Each dot indicates a box center.
    Color indicates size/shape as in~\cite{carion2020end}, i.e.~green represents small boxes while red/blue represents large horizontal/vertical boxes, respectively.
    Rows show the first three decoder slots (not cherry-picked).
    Columns show frames of synthetic videos created by sliding a cropped view over the image from left to right.
    Boxes with objectness \textgreater~0.2 are shown.
    On the first frame, like DETR, slots specialize to certain areas. However, on subsequent frames, they track appearance rather than remaining at the same image coordinates.
    The model learns to reserve some slots (e.g. Slot 2) for late-appearing objects.
    }    
    \vspace{-2mm}
    \label{fig:box-centers}
\end{figure}

\paragraph{Temporal association analysis}

\setlength{\intextsep}{3pt}%
\setlength{\columnsep}{3pt}%
\begin{wrapfigure}{r}{0.29\columnwidth}
    \vspace{-0.5em}
    \includegraphics[width=0.29\columnwidth]{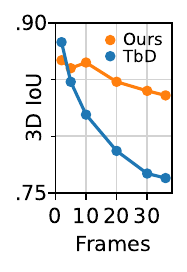} 
    \vspace{-2em}
    \caption{SOT.}
    \label{fig:single_object_tracking}
\end{wrapfigure}

To isolate temporal association (tracking) performance, we assess our method in the single-object-tracking (SOT) setting by initializing the track from the initial ground-truth bounding box on YT-VIS. We compare Video OWL-ViT to the tracking-by-detection (TbD) baseline in terms of 3D IoU ($\sum_t p_t \cap g_t / \sum_t p_t \cup g_t$), see Figure~\ref{fig:single_object_tracking}. 

Performance of the baseline decays faster than Video OWL-ViT performance, which indicates superior temporal association performance afforded by the end-to-end tracking architecture of Video OWL-ViT.

\paragraph{Location specificity of object queries}
To further illustrate the temporal association capability of Video OWL-ViT, we perform an analysis of the location specificity of object queries over time. In \Cref{fig:box-centers}, we visualize box predictions by object queries (slots). 

To disentangle whether slots attend to image space or object appearance, we create synthetic videos by sliding a cropped view over the image. We find that slots move with the image (i.e. with appearance), rather than remaining fixed at a specific location. Additionally, the model learns to reserve some slots for late-appearing objects.

\subsection{Understanding the Challenges of Open-World Video Modeling}

In the previous sections, we have presented an end-to-end trainable open-world video localization model.
We now analyze the error modes of our model and describe how we address them, with a particular focus on end-to-end learning of tracking dynamics from limited video data. Overall, we find that a realistic distribution of object dynamics in the training data, as well as careful modeling of object presence, are important factors affecting model performance.

\paragraph*{Training clip length} A common approach for video models~\cite{TrackFormer} is to train on the shortest possible clip length (i.e.\ two frames), which is memory and compute efficient. However, training on longer clips allows for more realistic learning of object dynamics, including motion, appearance/disappearance, occlusion, recovery from tracking errors, and long-term dependencies. This may be especially important for a model that learns object dynamics directly from the data.
We empirically confirm this in Table~\ref{table:tao-ow-clip-length}. We find that training on 4-frame clips significantly improves performance compared to training on 2-frame clips. Beyond 4 frames, performance quickly saturates.  %

\begin{table}[tbp]
\caption{Clip length for training on \textbf{TAO-OW} open world tracking (unknown classes).}
\vspace{-1em}
\label{table:tao-ow-clip-length}
\centering
\begin{tabular}{@{}lccc@{}} \\
\toprule
\textbf{Clip length} & OWTA$\uparrow$ & D.~Re.~$\uparrow$ & A.~Acc.~$\uparrow$  \\
\midrule
2 frames & 35.2 & 40.7 & 32.6\\
4 frames & 45.3 & 53.2 & 40.5\\
6 frames & 45.4 & 53.4 & 40.5\\
\bottomrule
\end{tabular}
\end{table}

\paragraph*{Training on pseudo-videos}
TAO-OW contains only 500 videos in the training set, which poses the risk of overfitting on the small number of represented object classes. To maintain performance on object classes not represented in the video data, we leverage more abundant image data by generating pseudo-videos from still images (\Cref{sec:video_owl_vit}). The image data includes the training data from LVIS~\cite{LVIS} (approx.~100k images) and Objects365~\cite{Objects365} (approx.~600k images). As shown in \Cref{table:tao-ow-supervision}, training on pseudo-videos yields a substantial improvement in OWTA on unknown classes compared to training on real videos. Performance is also improved for known classes, which is notable given that pseudo-videos do not have realistic motion dynamics and underscores the importance of sufficient training data.
Combining real and pseudo-videos further improves performance on known classes, but not on unknown classes. The fact that training on pseudo-videos alone performs similar on unknown classes compared to training on real videos illustrates how challenging the small amount of video data is for open-world performance. %
\begin{table}[tbp]
\caption{Comparison between fine-tuning OWL-ViT with pseudo-videos or real videos. Results on \textbf{TAO-OW} open world tracking; all metrics in $\%$.}
\vspace{-1em}
\label{table:tao-ow-supervision}
\centering
\resizebox{\linewidth}{!}{
\begin{tabular}{@{}lcc@{}} \\
\toprule
\textbf{Supervision} & \textbf{Known} OWTA  & \textbf{Unknown} OWTA \\
\midrule
Real videos only & 54.6  & 33.9 \\
Pseudo-videos only & 56.9 & 45.9 \\
Real + pseudo-videos & 59.0 & 45.4 \\
\bottomrule
\end{tabular}
}
\end{table}

\begin{table}[tbp]
  \caption{Score calibration and temporal mosaic data augmentation improve performance on medium and short video tracks. Results on \textbf{TAO-OW} open world tracking; all metrics in $\%$. The track length buckets are \textit{short} (shorter than 3 seconds), \textit{medium} (between 3 and 10 seconds), and \textit{long} (longer than 10 seconds), with the following distributions:
    11\%/25\%/64\% (known) and
    4\%/15\%/81\% (unknown).}
  \vspace{-1em}
  \label{table:tao-ow-short-tracks}
  \centering
  \resizebox{\linewidth}{!}{
  \addtolength{\tabcolsep}{-4pt}
  \begin{tabular}{@{}cc|cccccc@{}} \\
    \toprule
    Score & Temporal & \multicolumn{3}{c}{\textbf{Known} OWTA}  & \multicolumn{3}{c}{\textbf{Unknown} OWTA} \\
    calibration & mosaic
    & Short & Med. & Long & Short & Med. & Long \\
    \cmidrule(l{2pt}r{2pt}){1-5} \cmidrule(l{2pt}r{2pt}){6-8}
    $\times$                   & $\times$                 & 15.9 & 20.2   & 58.6  & 12.5 & 16.0   & 45.7  \\
    \checkmark                 & $\times$                 & 26.1 & 30.6   & 59.4  & 22.2 & 26.2   & 46.8  \\
    \checkmark                 & \checkmark               & 31.3 & 32.5   & 60.6  & 22.8 & 27.6   & 45.9  \\
    \bottomrule
  \end{tabular}
  }  %
\end{table}

\paragraph*{Performance on short tracks}
We find that association accuracy is significantly lower for short than for long tracks (\Cref{table:tao-ow-short-tracks}). 
One reason for lower performance on short tracks may be that the objectness score is not a well-calibrated indicator for deciding whether an instance is an object or background. %
Since short tracks contain more frames during which the object is not visible (i.e.\ ``background''), they are disproportionately affected by poor objectness calibration.
To mitigate this effect, we use a simple heuristic to mark parts of tracks as ``background'' (i.e.~no object) that significantly differ from the maximum objectness score $o_\mathrm{max}$ across the track. We find that a simple per-track threshold of $0.3\cdot o_\mathrm{max}$, below which we mark detections as background, suffices to significantly improve performance on shorter tracks (``Score calibration'' in \Cref{table:tao-ow-short-tracks}). 

A second reason for poor short-track performance may be that the training data contains fewer short than long tracks. To address this imbalance, we create artificially shortened tracks by concatenating short clips from different videos into longer sequences (``Temporal mosaic'' in \Cref{table:tao-ow-short-tracks}).

We find that both score calibration and video mosaic data augmentation can significantly improve performance on short and medium-length tracks. The effect of score calibration is especially large, despite using a simple heuristic for recalibration. This suggests that a more sophisticated learnable and directly supervised presence indicator may lead to further improvements.

\paragraph*{Inference frame rate}

While TAO is annotated at 1 FPS, the videos come at a higher frame rate. This allows us to operate the model at higher frame rates, using also intermediate frames to compute the predictions. For the metrics, only the predictions associated with annotated frames are kept.
According to Liu et al.~\cite{Liu2021OpeningUO}, using intermediate frames generally improves known accuracy, but harms unknown accuracy for the tracking-by-detection OWTB baseline. In contrast, Video OWL-ViT improves known accuracy without degrading unknown accuracy (Table~\ref{table:tao-ow-fps}), demonstrating the benefit of \textit{learning} frame-to-frame association from data instead of relying on a matching heuristic. This helps close the known-accuracy gap between OWTB and Video OWL-ViT (at the expense of increased compute).

\begin{table}[tbp]
\caption{Evaluation at different frame rates (FPS). Results on \textbf{TAO-OW} open world tracking; all metrics in $\%$.}
\vspace{-1em}
\label{table:tao-ow-fps}
\centering
\begin{tabular}{@{}ccc@{}} \\
\toprule
\textbf{FPS} & \textbf{Known} OWTA  & \textbf{Unknown} OWTA \\
\midrule
1 & 59.0  & 45.4 \\
2 & 59.0  & 45.4 \\
4 & 61.1 & 46.1 \\
8 & 60.5 & 45.5\\
\bottomrule
\end{tabular}
\end{table}
\section{Conclusion}
We introduced Video OWL-ViT, a simple end-to-end model for open-world localization and tracking in video. Video OWL-ViT builds on the open-world detection recipe of OWL-ViT %
and transfers an image-text pre-trained vision transformer model to video via fine-tuning and tracking-specific data augmentation. To enable temporally-consistent localization of objects, we add a decoder to OWL-ViT to decouple object queries from the input pixel grid and train using a tracking-aware set prediction loss.

Video OWL-ViT achieves performance competitive with tracking-by-detection baselines on the open-world TAO-OW benchmark, while presenting several advantages, such as matching-free tracking at test time and consistent performance even at higher frame rates for long videos. Our \mbox{analyses} of performance limitations of the model suggest that improving the amount and quality of video training data, and the modeling of object presence are promising future directions.

\section*{Acknowledgements}
We would like to thank Alexey Dosovitskiy for providing detailed feedback on an earlier draft of the paper.

{\small
\bibliographystyle{ieee_fullname}
\bibliography{references}
}

\clearpage
\appendix

\section{Model details}
\subsection{Segmentation head}
We train a segmentation head on top of the frozen OWL-ViT model solely to enforce the one-output-per-pixel constraint required by the OWTA metric. The segmentation head predicts cropped masks within the bounding boxes predicted by the main model. It consists of a ResNet-26 encoder and Hourglass mask heads as described in~\cite{birodkar2021surprising}, trained on Open~Images~V5~\cite{OpenImagesSegmentation,OpenImages}.

After training this head on the OWL-ViT model, we apply the same (frozen) head on object queries in Video OWL-ViT (without re-training or fine-tuning) to obtain rough segmentation masks.

\begin{figure}[bp]
    \centering
    \includegraphics[width=\linewidth,trim={0cm 0cm 0cm 0cm},clip]{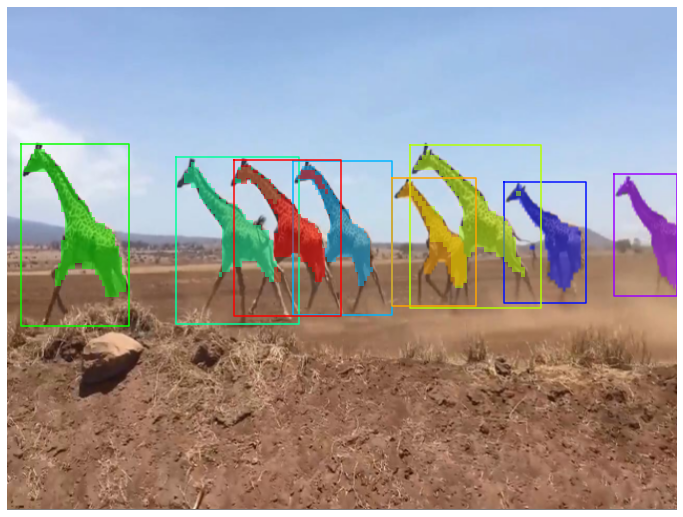}
    \caption{Example of segmentation masks used for enforcing the non-overlap constraint of the OWTA metric.
    \label{fig:segmentation-example}}
\end{figure}

Example qualitative segmentation masks are shown in Figure~\ref{fig:segmentation-example}.

\subsection{Architecture}
We provide an overview of architecture hyperparameters of Video OWL-ViT in Table~\ref{table:video_owl_vit_architecture}. We use pre-norm~\cite{xiong2020layer} in all transformer layers.

\begin{table}[h]
\caption{Video OWL-ViT architecture overview.}%
\vspace{-1em}
\label{table:video_owl_vit_architecture}
\centering
\begin{tabular}{@{}llr@{}} \\
\toprule
\textbf{Backbone} & \multicolumn{2}{c}{ViT-L/14} \\
\midrule
\multirow{6}{*}{\textbf{Decoder}} & Layers & 6 \\
 & Heads & 8 \\
 & Hidden dim & 1024 \\
 & MLP size & 4096 \\
 & QKV dim & 1024 \\
 & Dropout rate & 0.1 \\
\midrule
\multirow{2}{*}{\textbf{Box head}} & MLP size & 1024 \\
 & MLP hidden layers & 2 \\
 & MLP activation & GELU~\cite{hendrycks2016gaussian} \\
\bottomrule
\end{tabular}
\end{table}

\subsection{Data augmentation}
We use the following data augmentations for training on TAO-OW: 1) we randomly left-right flip all frames (jointly) in a training clip, 2) we randomly invert the temporal axis, 3) we apply random cropping (jointly across all frames in a clip), and 4) we apply a temporal video mosaic augmentation. All 6-frame clips used for training are randomly sampled from the training videos at 4FPS. 

For cropping, we sample a random $480\times640$ crop of the original video and discard bounding boxes if less than 50\% of their original box area remains after cropping. For temporal video mosaic, we take two processed video clips of length 6 (with augmentation as described above), concatenate them along the time axis, and sample a random temporal window of length 6 over the joint sequence. We apply temporal video mosaic to 50\% of training samples.

To obtain pseudo-videos from images (incl.~individual TAO-OW training frames), we apply a random crop (of size 50\% of height and width of the original image) that simulates linear camera motion over the image. We similarly discard bounding boxes if less than 50\% of their original box area remains after cropping.

\subsection{Training}
We train Video OWL-ViT using the Adam~\cite{kingma2014adam} optimizer with $\beta_1=0.9$, $\beta_2=0.999$, and with a batch size of 32 and a learning rate of 3e-6 for 100k training steps. We clip gradients to a maximum norm of 1. We linearly ``warm up'' the learning rate over the first 1k steps and decay it to 0 over the course of training using a cosine schedule. 

For our loss, we use the same hyperparameters as OWL-ViT~\cite{minderer2022simple}, i.e.~equal weighting between bounding box, gIoU, and classification losses, and focal loss coefficients of $\alpha=0.3$ and $\gamma=2$.

For simplicity, we do not filter class labels in upstream text-image and detection pre-training, i.e.~objects of classes that are considered "unknown" in the TAO-OW video tracking setting can appear in static images during training, but are never seen in natural video. We verified that filtering these classes during upstream pre-training has negligible effect on our reported metrics.

\section{Additional results}

\subsection{Backbone size}
To evaluate the effect of model size, we compare our default Video OWL-ViT model, which uses a ViT-L/14 backbone, to a model variant with a smaller backbone (ViT-B/16 at $768\times768$ resolution). Our results in Table~\ref{table:tao-ow-backbone-size} indicate clear performance gains when using the larger ViT-L/14 backbone across all metrics, incl.~upstream LVIS detection performance.

\begin{table}[t]
\caption{\textbf{TAO} open world tracking with Video OWL-ViT for different ViT backbone size. All metrics in $\%$.}%
\vspace{-1em}
\label{table:tao-ow-backbone-size}
\centering
\resizebox{\linewidth}{!}{
\addtolength{\tabcolsep}{-2pt}
\begin{tabular}{@{}llcccccccc@{}} \\
\toprule
 & & \multicolumn{2}{c}{\textbf{LVIS}}  & \multicolumn{3}{c}{\textbf{Known}}  & \multicolumn{3}{c}{\textbf{Unknown}} \\
 \cmidrule(l{2pt}r{2pt}){3-4}
 \cmidrule(l{2pt}r{2pt}){5-7} \cmidrule(l{2pt}r{2pt}){8-10}
\textbf{ViT} & \textbf{Resolution} & AP & APr & OWTA & D.~Re. & A.~Acc. & OWTA & D.~Re. & A.~Acc. \\
\cmidrule(l{2pt}r{2pt}){1-4}
\cmidrule(l{2pt}r{2pt}){5-7} \cmidrule(l{2pt}r{2pt}){8-10}
B/16 & 768 & 27.2 & 20.6 & 55.2 & 64.3 & 48.9 & 41.6 & 48.6 & 37.9\\
L/14 & 672 & 33.4 & 31.8 & 59.0 & 69.0 & 51.5 & 45.4 & 53.4 & 40.5\\
\bottomrule
\end{tabular}
}
\end{table}

\subsection{Qualitative results}
\begin{figure*}[tbp]
    \centering
    \begin{subfigure}{1.00\linewidth}
    \includegraphics[width=\linewidth,trim={0cm 0cm 364cm 0cm},clip]{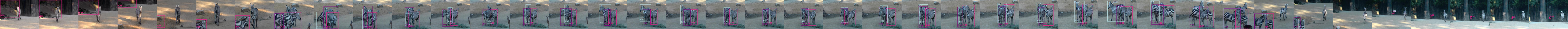}
    \end{subfigure}
    \begin{subfigure}{1.00\linewidth}
    \includegraphics[width=\linewidth,trim={0cm 0cm 364cm 0cm},clip]{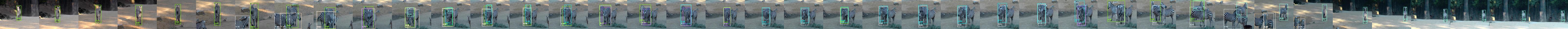}
    \end{subfigure}
    \begin{subfigure}{1.00\linewidth}
    \includegraphics[width=\linewidth,trim={0cm 0cm 364cm 0cm},clip]{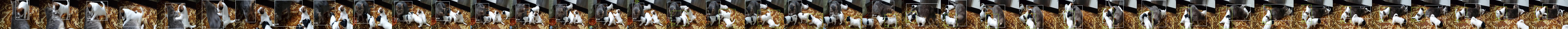}
    \end{subfigure}
    \begin{subfigure}{1.00\linewidth}
    \includegraphics[width=\linewidth,trim={0cm 0cm 364cm 0cm},clip]{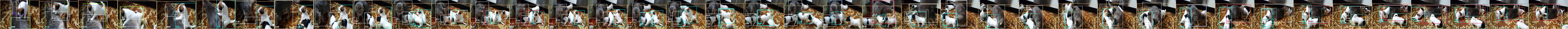}
    \end{subfigure}
    \begin{subfigure}{1.00\linewidth}
    \includegraphics[width=\linewidth,trim={0cm 0cm 364cm 0cm},clip]{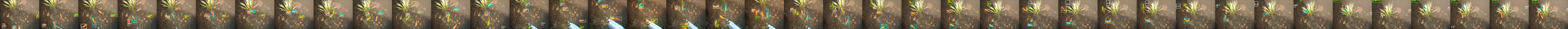}
    \end{subfigure}
    \begin{subfigure}{1.00\linewidth}
    \includegraphics[width=\linewidth,trim={0cm 0cm 364cm 0cm},clip]{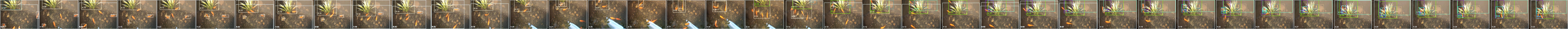}
    \end{subfigure}
    \begin{subfigure}{1.00\linewidth}
    \includegraphics[width=\linewidth,trim={0cm 0cm 364cm 0cm},clip]{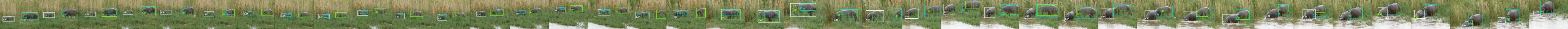}
    \end{subfigure}
    \begin{subfigure}{1.00\linewidth}
    \includegraphics[width=\linewidth,trim={0cm 0cm 364cm 0cm},clip]{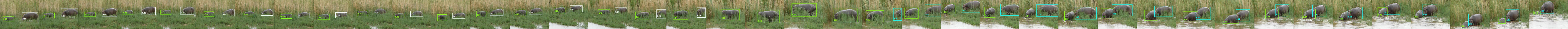}
    \end{subfigure}
    \begin{subfigure}{1.00\linewidth}
    \includegraphics[width=\linewidth,trim={0cm 0cm 364cm 0cm},clip]{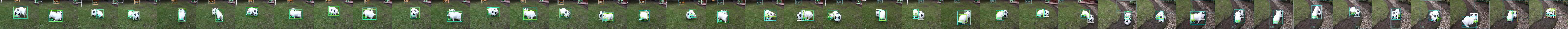}
    \end{subfigure}
    \begin{subfigure}{1.00\linewidth}
    \includegraphics[width=\linewidth,trim={0cm 0cm 364cm 0cm},clip]{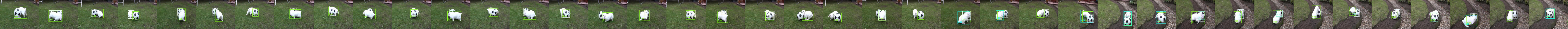}
    \end{subfigure}
    \caption{Qualitative examples for Video OWL-ViT detection and tracking of multiple instances on the \textbf{TAO-OW} validation set. Tracking-by-detection (odd rows) vs Video OWL-ViT (even rows). 
    Known classes include: cat, dog, zebra.
    Unknown classes include: fish, rabbit, hippopotamus.
    Colors uniquely correspond to query IDs. Numbers indicate objectness scores. 
    Only the first 6 frames/seconds of each video are shown.
    \label{fig:tao-ow-qualitative-results}}
\end{figure*}

We show further qualitative results of high scoring tracks for Video OWL-ViT and our tracking-by-detection baseline in Figure~\ref{fig:tao-ow-qualitative-results} (TAO-OW) and Figure~\ref{fig:yt-vis-ow-qualitative-results} (YT-VIS). Qualitative results in video format are provided in the supplementary zip file. Video OWL-ViT generally maintains consistent tracks and avoids transfer of instance predictions across semantically different objects compared to our tracking-by-detection baseline.

\begin{figure*}[tbp]
    \centering
    \begin{subfigure}{0.64\linewidth}
    \includegraphics[width=\linewidth,trim={0cm 0cm 0cm 0cm},clip]{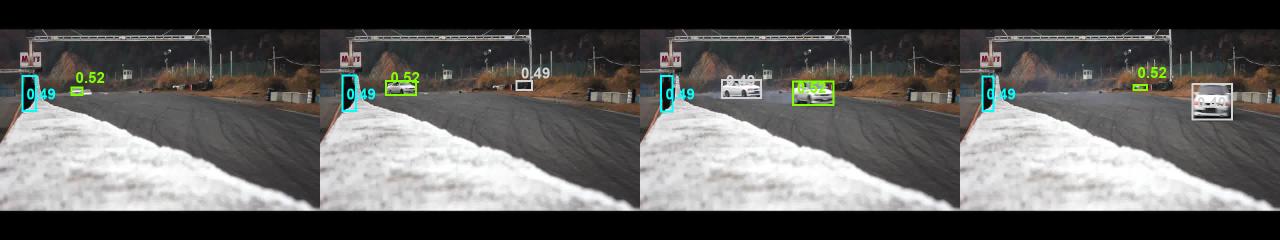}
    \end{subfigure}$\phantom{EMPTY_FRAME}\phantom{EMPTY_FRAME}$\\
    \begin{subfigure}{0.64\linewidth}
    \includegraphics[width=\linewidth,trim={0cm 0cm 0cm 0cm},clip]{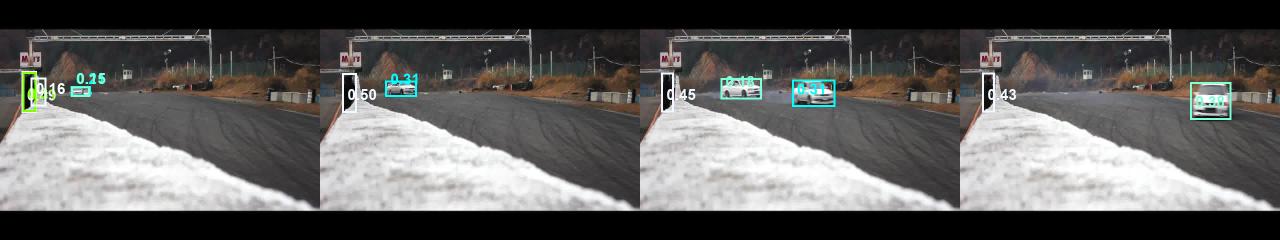}
    \end{subfigure}$\phantom{EMPTY_FRAME}\phantom{EMPTY_FRAME}$\\
    \begin{subfigure}{0.48\linewidth}
    \includegraphics[width=\linewidth,trim={0cm 0cm 0cm 0cm},clip]{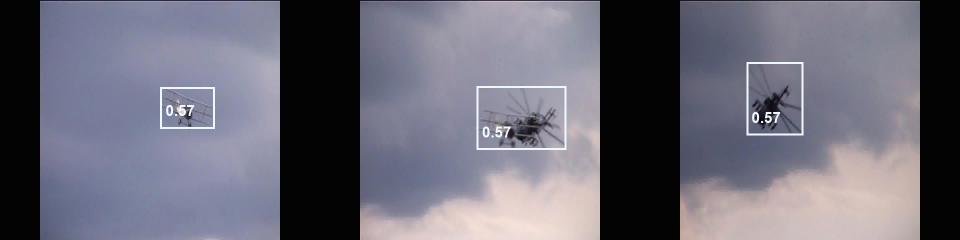}
    \end{subfigure}$\phantom{EMPTY_FRAME}\phantom{EMPTY_FRAME}\phantom{EMPTY_FRAME}$\\
    \begin{subfigure}{0.48\linewidth}
    \includegraphics[width=\linewidth,trim={0cm 0cm 0cm 0cm},clip]{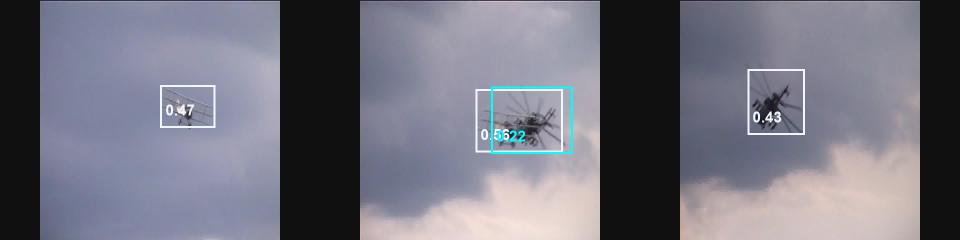}
    \end{subfigure}$\phantom{EMPTY_FRAME}\phantom{EMPTY_FRAME}\phantom{EMPTY_FRAME}$\\
    \begin{subfigure}{0.64\linewidth}
    \includegraphics[width=\linewidth,trim={0cm 0cm 0cm 0cm},clip]{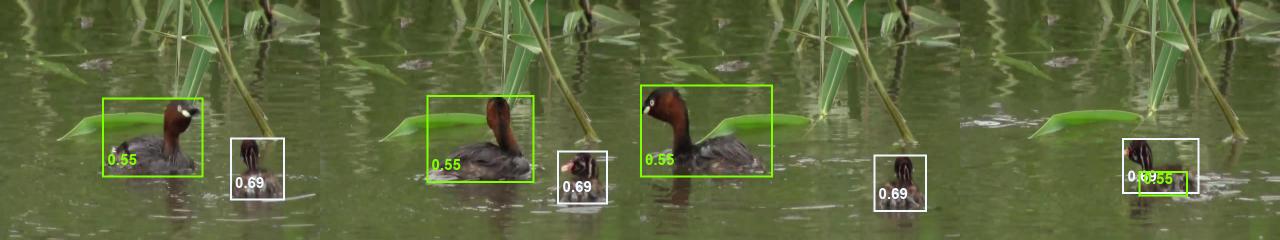}
    \end{subfigure}$\phantom{EMPTY_FRAME}\phantom{EMPTY_FRAME}$\\
    \begin{subfigure}{0.64\linewidth}
    \includegraphics[width=\linewidth,trim={0cm 0cm 0cm 0cm},clip]{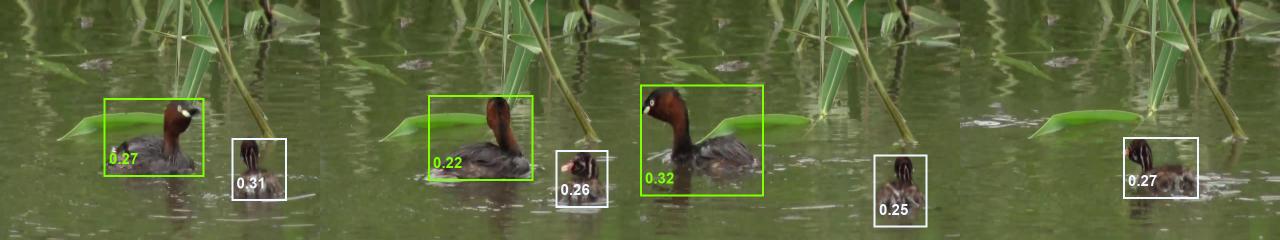}
    \end{subfigure}$\phantom{EMPTY_FRAME}\phantom{EMPTY_FRAME}$\\
    \begin{subfigure}{0.96\linewidth}
    \includegraphics[width=\linewidth,trim={0cm 0cm 0cm 0cm},clip]{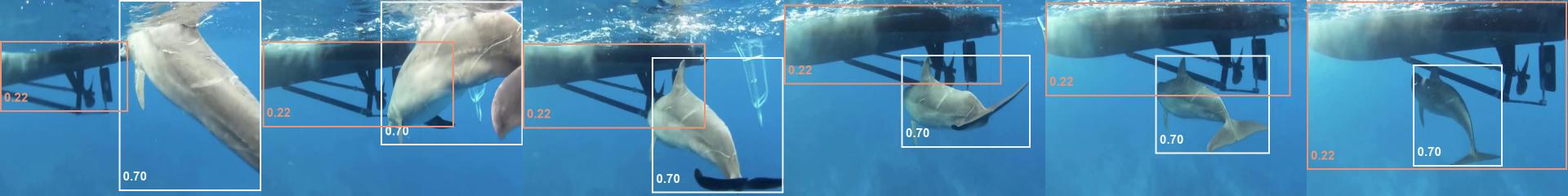}
    \end{subfigure}\\
    \begin{subfigure}{0.96\linewidth}
    \includegraphics[width=\linewidth,trim={0cm 0cm 0cm 0cm},clip]{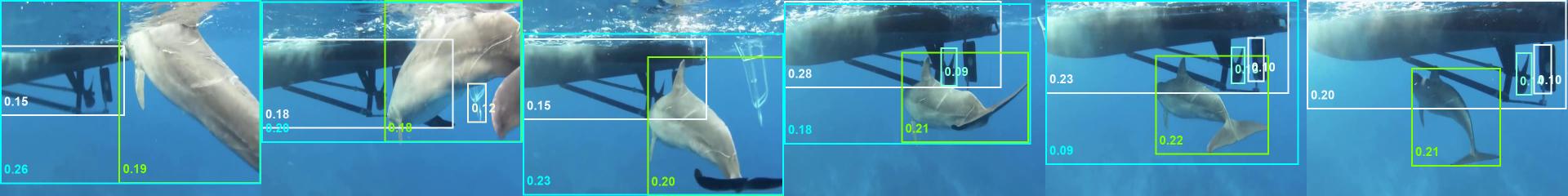}
    \end{subfigure}\\
    \caption{Qualitative examples for Video OWL-ViT detection and tracking of multiple instances on the \textbf{YT-VIS} validation/test sets. Tracking-by-detection (odd rows) vs Video OWL-ViT (even rows). 
    Known classes include: dog, car, airplane.
    Unknown classes include: duck, shark. 
    Colors uniquely correspond to query IDs. Numbers indicate objectness scores. 
    The video clips are shown at a reduced frame rate (1 FPS).
    \label{fig:yt-vis-ow-qualitative-results}}
\end{figure*}

\end{document}